\documentclass[lettersize,journal]{IEEEtran}
\usepackage{amsmath,amsfonts}
\usepackage{algorithm}
\usepackage{array}
\usepackage[caption=false,font=normalsize,labelfont=sf,textfont=sf]{subfig}
\usepackage{textcomp}
\usepackage{stfloats}
\usepackage{url}
\usepackage{verbatim}
\usepackage{graphicx}
\usepackage{cite}
\usepackage{xcolor}
\usepackage{amsmath}
\usepackage{amssymb}  
\usepackage{enumitem}
\usepackage{algpseudocode}
\usepackage{array}
\usepackage{tabularx}
\usepackage{booktabs}
\usepackage{multirow} 
\usepackage{graphicx}
\usepackage{adjustbox}
\usepackage[table]{xcolor} 
\usepackage{url} 
\definecolor{lightgray}{RGB}{240,240,240} 

\usepackage{pifont}
\newcommand{\cmark}{\ding{51}} 
\newcommand{\xmark}{\ding{55}}
\usepackage{tcolorbox}
\usepackage{fontawesome} 
\definecolor{bggray}{RGB}{245,245,245} 
\definecolor{codeblue}{RGB}{0,0,139}   
\usepackage[colorlinks=true, urlcolor=blue, linkcolor=black, citecolor=black]{hyperref}
\definecolor{codegreen}{RGB}{0,100,0}
\usepackage{booktabs}
\usepackage{multirow}
\usepackage{tabularx}
\usepackage[table]{xcolor}

\begin{document}

\title{HCSG: Human-Centric Semantic-Geometric Reasoning for Vision-Language Navigation}
\author{Haoxuan~Xu,
        Tianfu~Li,
        Wenbo~Chen,
        Yi~Liu,
        Jin~Wu,
        Huashuo~Lei,
        Yunfan~Lou,
        Lujia~Wang,
        Hesheng~Wang,
        Haoang~Li%
\thanks{Haoxuan Xu, Tianfu Li, Wenbo Chen, Huashuo Lei, Lujia Wang and Haoang Li are with The Hong Kong University of Science and Technology (Guangzhou), Guangzhou 511453, China (e-mail: hxu095@connect.hkust-gz.edu.cn; tli794@connect.hkust-gz.edu.cn; wchen361@connect.hkust-gz.edu.cn; hlei573@connect.hkust-gz.edu.cn; eewanglj@connect.hkust-gz.edu.cn; haoang.li.cuhk@gmail.com). (Haoxuan~Xu
and Tianfu Li contributed equally to this work.) (Corresponding
author: Haoang Li.)}%
\thanks{Yi Liu is with Tsinghua University, Shenzhen 518055, China (e-mail: yiliu24@mails.tsinghua.edu.cn).}%
\thanks{Jin Wu is with University of Science and Technology Beijing, Beijing
100083, China (e-mail: wujin@ustb.edu.cn).}%
\thanks{Yunfan Lou is with National University of Singapore,
119077, Singapore (e-mail: e1373933@u.nus.edu).}%
\thanks{Hesheng Wang is with Shanghai Jiao Tong University, Shanghai 200240, China (e-mail: wanghesheng@sjtu.edu.cn).}%
}


\maketitle

\begin{abstract}

VLN has achieved remarkable progress by scaling data and model capacity. However, the assumption of a static environment breaks down in real-world indoor scenarios, where robots inevitably encounter dynamic pedestrians. Existing human-aware approaches typically treat humans merely as moving obstacles based on implicit visual cues, lacking the explicit reasoning required to interpret human intentions or maintain social norms. To address this, we propose HCSG, the first human-centric framework for VLN. This framework provides a robust foundation for safe, socially intelligent navigation in dynamic human-robot environments that shifts the paradigm from passive collision avoidance to active human behavior understanding. Specifically, HCSG introduces a unified Human Understanding Module that synergizes two key capabilities: (i) geometric forecasting, which predicts human pose and trajectory to anticipate future motion dynamics; and (ii) semantic interpretation, which leverages a Vision-Language Model (VLM) to generate natural language descriptions of human actions and intentions. These semantic-geometric representations are fused into the agent’s topological map for instruction-conditioned planning. Furthermore, a social distance loss is introduced to enforce socially compliant interaction distances. Extensive experiments on the HA-VLNCE benchmark demonstrate that HCSG significantly outperforms state-of-the-art methods, achieving a 14\% improvement in Success Rate and a 34\% reduction in Collision Rate. Our project can be seen at \url{https://haoxuanxu1024.github.io/HCSG/}.
\end{abstract}


\section{Introduction}

Vision-Language Navigation (VLN)~\cite{an2022bevbert,an2024etpnav,anderson2018vision,ku2020room} enables robots to follow multimodal instructions and navigate physical spaces. Given a language instruction, an embodied agent must interpret commands together with egocentric visual observations to reach a target location or object. Recent advances have propelled VLN from simplified discrete settings~\cite{ku2020room} to complex continuous environments~\cite{an2024etpnav,krantz2020beyond}, increasingly leveraging large-scale pre-training and foundation models to improve instruction grounding and embodied reasoning~\cite{chen2024mapgpt,su2024boost,zhou2024navgpt}. As such, VLN has become a key capability for service robots in indoor environments.

\begin{figure}[!t]
    \centering
    \includegraphics[width=1\linewidth]{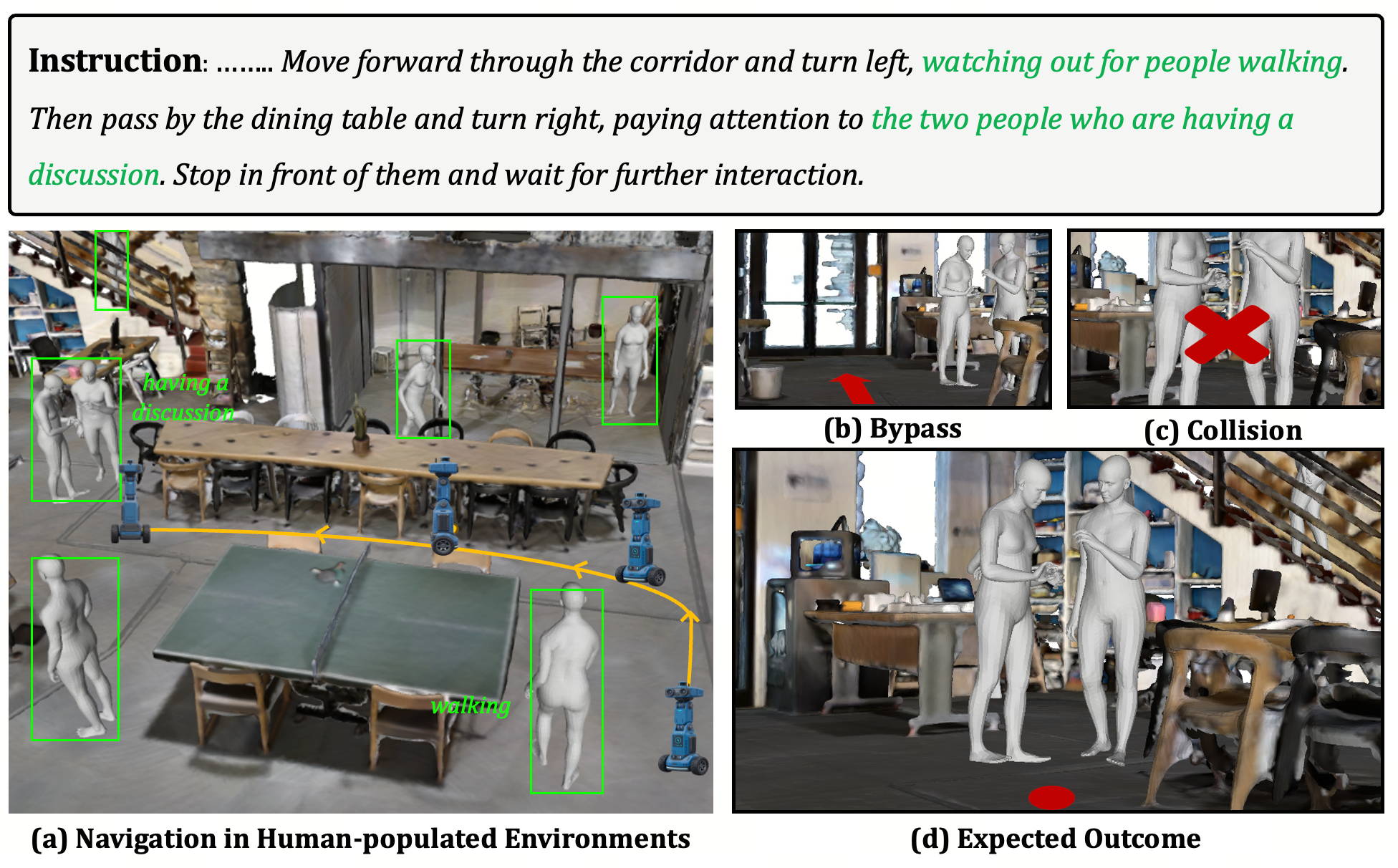}
    \caption{An illustrative navigation scenario in a human-populated environment. The input is an instruction ``\textit{pay attention to the two people who are having a discussion. Stop in front of them...}''. (a) \textbf{Overview}: A typical scenario demonstrating the challenge of dynamic humans. (b) \textbf{Bypass}: Without semantic understanding, traditional agents may misclassify task-relevant people as generic obstacles and avoid them entirely. (c) \textbf{Collision}: Without geometric understanding, traditional agents often fail to anticipate future motion, leading to physical collisions. (d) \textbf{Expected Outcome}: Our HCSG synergizes semantic interpretation and geometric forecasting to achieve instruction-grounded and socially compliant navigation.}
    \label{fig:intro}
\end{figure}

However, the prevailing VLN paradigm largely assumes a static environment~\cite{anderson2018vision}, which breaks down in real-world settings where robots inevitably encounter moving pedestrians. Safe navigation requires anticipating future
pedestrian motion, a problem widely studied in trajectory prediction~\cite{quan2021holistic}. To narrow this realism gap, recent works have augmented simulators with human-populated scenes~\cite{dong2025ha,szot2021habitat,habitat19iccv,puig2023habitat3,krantz2020beyond}. Yet existing VLN agents still treat pedestrians primarily as moving obstacles rather than task-relevant entities. Their representations rely mainly on global visual cues and lack explicit modeling of human activities, intentions, and social interaction signals. Although several pioneering efforts have introduced partial forms of human awareness, they either restrict interaction to explicit dialogue~\cite{liu2024dragon,wang2025cori} or use vision-language models mainly for passive collision avoidance~\cite{payandeh2024social,song2024vlm}. As a result, current navigation policies lack explicit reasoning about what humans are doing, where they are moving, and how the robot should behave around them under social norms. This limitation leads to severe failures in dynamic human-populated environments, as illustrated in Fig.~\ref{fig:intro}. Consider the instruction \textit{``pay attention to the two people... having a discussion
and stop in front of them''}. A conventional agent may bypass the target group entirely because it cannot semantically recognize their ongoing activity and thus misclassifies them as obstacles (Fig.~\ref{fig:intro}(b)). Conversely, even if it approaches the group, it may fail to anticipate their personal space and future motion, resulting in an intrusive or even colliding trajectory (Fig.~\ref{fig:intro}(c)). These failures reveal a fundamental gap in current VLN: the lack of explicit human-centric reasoning, which requires both semantic understanding of human activities and geometric foresight of human motion.

To address this gap, we propose \textbf{HCSG}, a \textbf{H}uman-\textbf{C}entric \textbf{S}emantic-\textbf{G}eometric Reasoning framework for VLN that explicitly equips the agent with semantic understanding and geometric foresight. The central idea is that successful navigation in human-populated environments requires not only recognizing where people are, but also reasoning about what they are doing, how they may move, and how the robot should behave with respect to them under the language goal. Accordingly, HCSG departs from conventional VLN agents that react to single-frame observations or rank navigable candidates primarily from static environmental cues~\cite{an2024etpnav,an2022bevbert,9879946}. Instead, once humans are detected, the agent pauses briefly to collect a short temporal observation window, providing dynamic context for subsequent human-aware reasoning.

Based on this observation sequence, HCSG processes human information through two complementary streams. To address the semantic deficit illustrated in Fig.~\ref{fig:intro}(b), the \textbf{Semantic Stream} leverages a large vision-language model (VLM) to analyze cropped human-centered observations and produce explicit descriptions of actions, intentions, and social context. These semantic representations enable the agent to move beyond treating people as undifferentiated obstacles, and instead align observed human behavior with the instruction, for example by identifying \textit{a group of people discussing} as the intended interaction target. To address the geometric deficit illustrated in Fig.~\ref{fig:intro} (c), the \textbf{Geometric Stream} models human dynamics through fine-grained pose cues and future trajectory forecasting. By encoding short-term temporal observations, this stream captures spatio-temporal patterns of motion and anticipates future human occupancy, providing predictive signals for proactive collision avoidance. Moreover, pose structure offers complementary physical evidence for behavior understanding, such as body orientation and coordinated group activity, thereby grounding the semantic interpretation in observable motion patterns.

The resulting semantic and geometric cues are fused into the agent's topological representation, so that human-aware states become explicitly available to the navigation policy rather than remaining buried in global scene features. This enriched representation allows a cross-modal transformer to align the global linguistic goal with localized human-centric nodes and select waypoints that are not only goal-directed, but also socially appropriate. To further enforce safe interaction, we introduce a \textbf{Social Distance Loss} that penalizes trajectories violating predicted human occupancy and interpersonal space. This objective combines strict collision avoidance with softer repulsion from future human regions, encouraging the agent to respect socially comfortable distances during approach and navigation. In this way, HCSG moves beyond passive obstacle avoidance toward instruction-conditioned human-centric navigation in dynamic environments.

Extensive experiments on the HA-VLNCE benchmark~\cite{dong2025ha} demonstrate that HCSG achieves state-of-the-art performance, improving Success Rate by 14.3\% and reducing Collision Rate by 34.5\% on the challenging validation-unseen split.

In summary, the contributions of this work are four-fold:

\begin{itemize}
    \item We formulate human-centric VLN as a navigation problem that requires both semantic understanding of human activities and geometric foresight of human motion, moving beyond the conventional view of pedestrians as mere obstacles.
    \item We propose HCSG, a dual-stream human-aware reasoning framework that combines VLM-based semantic interpretation with pose- and trajectory-based motion modeling for instruction-conditioned navigation in dynamic social environments.
    \item We introduce a Social Distance Loss to discourage trajectories that violate predicted human occupancy and interpersonal space, promoting safe and socially compliant navigation.
    \item HCSG achieves state-of-the-art performance on the HA-VLNCE benchmark, especially in success rate and collision rate.
\end{itemize}

\section{Related Work}
\subsection{Vision-Language Navigation}
VLN requires an embodied agent to interpret natural language instructions together with egocentric visual observations in order to reach a goal in either discrete~\cite{anderson2018vision} or continuous environments~\cite{krantz2020beyond}. Early work relied on sequence‑to‑sequence architectures, including recurrent neural network (RNN)~\cite{an2021neighbor,dang2022unbiased,he2023mlanet}, Long Short Term Memory network (LSTM)~\cite{anderson2018vision}, and Transformer‑based models~\cite{hong2021vln}. Subsequent advances improved VLN along several axes: more effective learning strategies~\cite{wang2019reinforced,li2026p}, data‑augmentation techniques~\cite{li2022envedit,wang2023scaling,fried2018speaker,tan2019learning,xu2026enhancing}, large‑scale pre‑training~\cite{guhur2021airbert,hao2020towards,huang2019transferable,majumdar2020improving,qiao2023hop+}, LLM-based methods~\cite{chen2024mapgpt,zhou2024navgpt} instruction-aware semantic enhancement~\cite{dai2026thinkmatter} and language-driven spatial localization~\cite{shi2025langloc}. To strengthen spatial reasoning, DUET~\cite{chen2022think} and ETPNav~\cite{an2024etpnav} employed topological maps that record the agent’s trajectory for global planning, while BEVBert~\cite{an2022bevbert} and GridMM~\cite{wang2023gridmm} generated bird’s‑eye‑view (BEV) grid maps for geometry‑consistent scene understanding; VER~\cite{liu2024volumetric} further extended this idea to 3D voxel grids. However, these methods largely focus on scene understanding in static environments and provide limited support for reasoning about dynamic human activity, especially in indoor spaces where pedestrians may appear, move, and interact unpredictably.

\subsection{Human-aware Understanding}
Once navigation moves beyond static scenes, a key challenge is no longer only where to go, but also how to interpret the humans encountered along the way. Human-aware understanding therefore ranges from fine-grained human attribute grounding~\cite{niu2024comprehensive} to modeling human motion, actions, and intentions.
Trajectory-centric pedestrian models predict motion and intent~\cite{Wang_2024_CVPR}, key-point pose predictors encode skeletal dynamics~\cite{delmas2022posescript}, and large-scale foundation models further distill the semantics of continuous human motion~\cite{chen2024motionllm}, together supporting reasoning about dynamic human-environment interactions.
Yet within the VLN literature, current policies still make limited use of human-aware cues for modeling actions and intentions.
Projects such as DRAGON~\cite{liu2024dragon} and CoRI~\cite{wang2025cori} let robots infer user goals indirectly through dialogue, and Social‑LLaVA~\cite{payandeh2024social} and VLM‑Social‑Nav~\cite{song2024vlm} deploy vision‑language models to detect pedestrian intent for collision avoidance; however, these efforts remain focused on simply steering clear of people. 
Genuine human awareness entered VLN research only with the release of the HA‑VLN~\cite{li2024human} and HA‑VLNCE~\cite{dong2025ha} datasets, but their baseline agents still view pedestrians as moving obstacles and lack deeper insight into human actions and intentions. 
Therefore, the next step for VLN is to move beyond treating humans as mere moving obstacles, toward explicitly understanding human actions and intentions so that agents can ground instructions more accurately in dynamic human-populated environments.

\begin{figure*}[t!]
    \centering
    \includegraphics[width=1\linewidth]{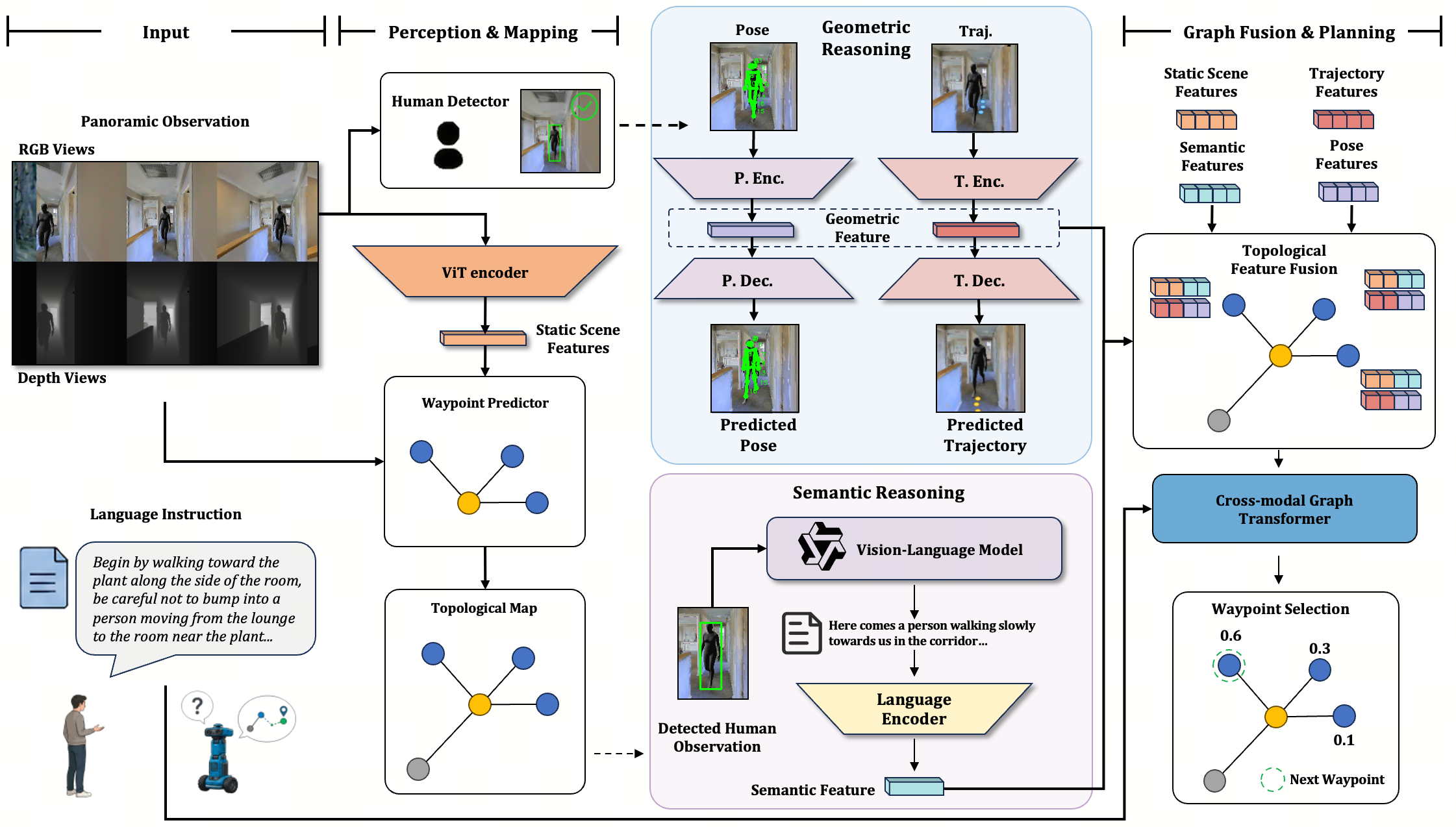}
    \caption{Overview of the proposed Human-Centric Semantic-Geometric Reasoning framework for Vision-Language Navigation (HCSG). Starting from panoramic observations and a language instruction, the agent performs perception and mapping, followed by parallel \textbf{Geometric Reasoning} and \textbf{Semantic Reasoning} for detected humans. The resulting human-centric features are fused with static scene features in the topological graph, which is then processed by a Cross-modal Graph Transformer for language-guided planning, i.e. waypoint selection.}
    \label{fig:pipeline}
\end{figure*}

\subsection{Social Robot Navigation}
Beyond understanding human behavior, robots must also respond to it in socially appropriate ways. Social robot navigation therefore extends obstacle avoidance by requiring agents to follow implicit social norms, such as maintaining comfortable interpersonal distances and producing legible motion.
From a control perspective, geometric and optimization-based approaches model social compliance through hybrid navigation architectures, human-aware costs, elastic-band refinement, and short-horizon pedestrian motion prediction~\cite{zhang2024decouple,jang2024socialzone,perez2024socialelastic,nguyen2024integrated,li2026scsv}.
Other studies further emphasize legibility and predictability as key properties of socially appropriate motion in multi-agent environments~\cite{bilen2024adaptiveproxemics}.
From a learning perspective, Deep Reinforcement Learning and risk-aware Model Predictive Control have been used to model complex multi-agent interactions under predictive uncertainty~\cite{xue2024crowdaware,sun2025lrmcp,gong2024falcon}. 
Despite these advances, most social navigation systems remain task-agnostic, treating pedestrians as generic dynamic obstacles to be avoided or politely bypassed. Yet social compliance alone is insufficient for VLN. While human-aware understanding provides cues about human actions and intentions, and social navigation provides principles for socially appropriate response, neither alone supports instruction-conditioned human-centric navigation. An agent must understand who a person is in relation to the language goal, what that person is doing, and how to navigate around or toward them appropriately. This gap motivates our HCSG framework.

\section{Problem Formulation}

Our work mainly follows the setup of Vision-Language Navigation in Continuous Environments (VLN-CE).
For most VLN-CE works~\cite{an2024etpnav,an2022bevbert,wang2023gridmm}, the task requires an embodied agent $\mathcal{A}$ to navigate in 3D environments $\mathcal{E}$, guided by natural language instruction $\mathcal{I}$. Formally, at each timestep $t \in [0, T]$:
$\mathcal{A}$ receives spherical observation $\mathbb{O}_t = (\mathcal{V}_t^{\mathrm{rgb}}, \mathcal{V}_t^{\mathrm{depth}})$ with 
$\mathcal{V}_t^{\mathrm{rgb}} = \{ v_{t,k}^{\mathrm{rgb}} \}_{k=1}^{12}$,
$\mathcal{V}_t^{\mathrm{depth}} = \{ v_{t,k}^{\mathrm{depth}} \}_{k=1}^{12}$ 
covering 360\textdegree~FoV at $30^\circ$ intervals. 
The agent learns a policy $\pi : (\mathbb{O}_t, \mathcal{I}) \rightarrow \mathbf{a}_{t+1}$ 
which outputs $\mathbf{a}_{t+1} \in \mathbb{R}^3$ representing relative displacement $(\Delta x, \Delta y, \Delta\theta)$ in $\mathrm{SE}(2)$ space.
The navigation trajectory $\tau = \{ \mathbf{p}_t \}_{t=0}^T$ terminates when $\| \mathbf{p}_T - \mathbf{p}_{\mathrm{goal}} \|_2 < \delta_{\mathrm{th}}$. Additionally, in our human-aware VLN-CE task setting, the observations $\mathbb{O}_t$ of the agent include dynamic humans, and the instructions $\mathcal{I}$ also contain descriptions of people in the scene.


\section{Methodology}
Based on the standard waypoint-based navigation policy~\cite{an2024etpnav,an2022bevbert,9879946}, we augment the conventional VLN framework with a human-centric reasoning pipeline. As illustrated in Fig.~\ref{fig:pipeline}, the overall navigation process first builds an online topological graph from panoramic observations and candidate waypoints, and then injects human-centric features into the graph whenever humans are detected. We describe this integration mechanism in Sec.~\ref{sec:Human Reasoning}.

Once a human is detected at a waypoint, the agent activates two complementary reasoning streams. The Geometric Reasoning Module, detailed in Sec.~\ref{sec:Geometric Reasoning}, estimates observed human pose and trajectory and further forecasts their future evolution. The Semantic Reasoning Module, detailed in Sec.~\ref{sec:Semantic Reasoning}, leverages a VLM to infer human actions, intentions, and navigation-relevant social context. The resulting semantic-geometric features are fused into the online topological representation for language-guided waypoint selection. Finally, the Social Distance Loss in Sec.~\ref{sec:Social Distance Loss} regularizes the navigation policy toward collision-free and socially compliant behavior.

\begin{figure}[t!]
    \centering
    \includegraphics[width=1\linewidth]{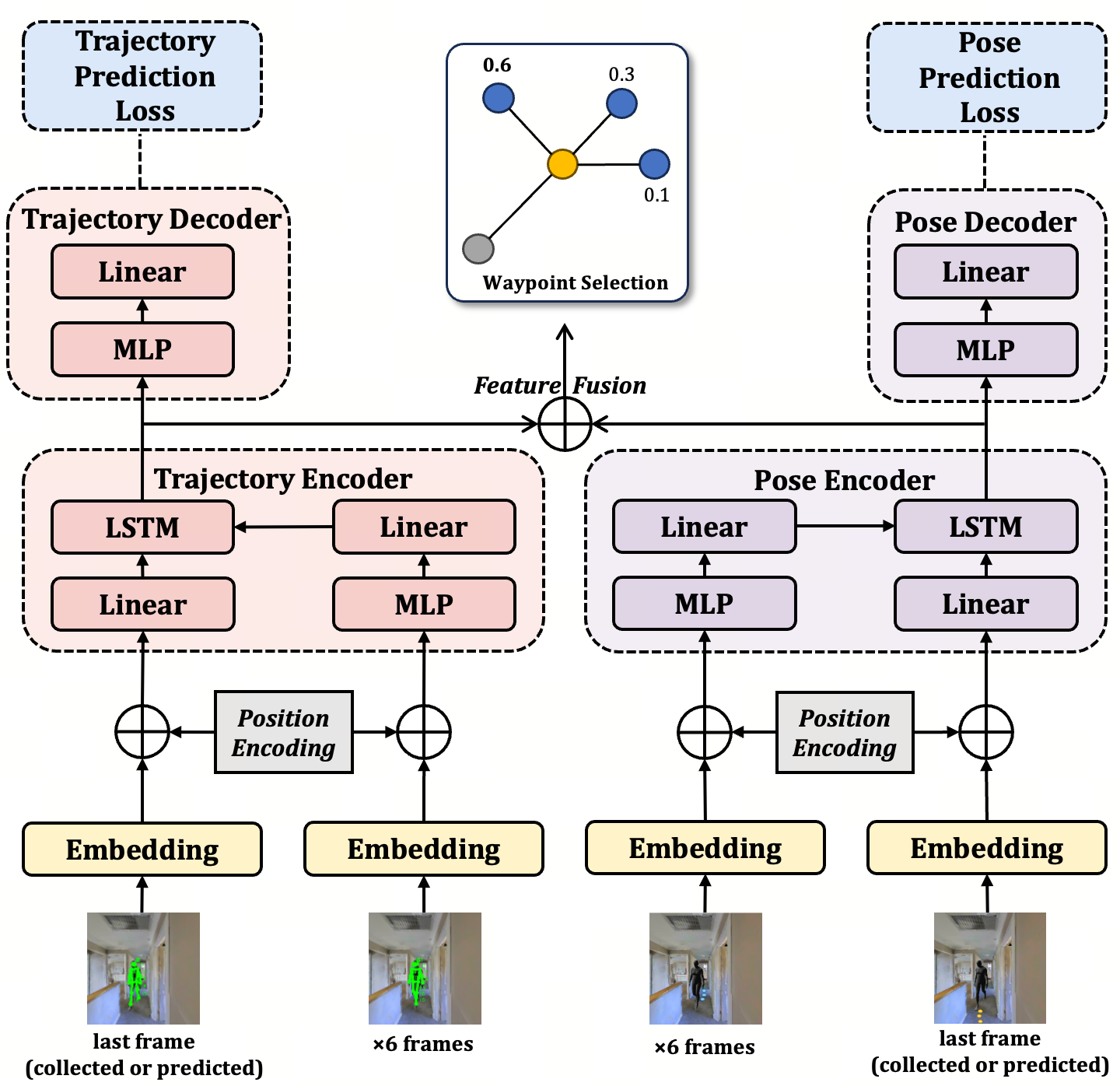}
    \caption{Pipeline of the Human Geometric Reasoning Module. We decompose the geometric motion of the human body into two complementary components: trajectory and pose. Both the trajectory and pose predictors utilize an LSTM-based encoder-decoder structure to process the input frame embeddings. By predicting these components and fusing intermediate features extracted from the network, we obtain future-oriented geometric representations. These fused features are subsequently supplied to the downstream navigation module for tasks such as waypoint selection.}
    \label{fig:geo}
\end{figure}

\subsection{Navigation with Human Reasoning}
\label{sec:Human Reasoning}
In this subsection, we describe how human-centric reasoning is integrated into the standard waypoint-based VLN pipeline. As illustrated in Fig.~\ref{fig:pipeline}, our method injects human features into the topological graph during navigation upon human detection. We next detail the concrete mechanism for integrating human reasoning into the navigation process.

\textbf{Waypoint-based Navigation.} At timestep $t$, the agent receives a panoramic observation $\mathbb{O}_t = (\mathcal{V}_t^{\mathrm{rgb}}, \mathcal{V}_t^{\mathrm{depth}})$ and passes it to an external waypoint predictor $f_{\text{way}}$ \cite{an2024etpnav} to estimate navigable positions, generating candidate nodes $\mathcal{W}_{t}$ for the next action. Meanwhile, the pre-trained visual encoder extracts features from each perspective and fuses them into the corresponding nodes estimated by the waypoint predictor:
\begin{equation}
\mathcal{F}_{t,k}^{\mathrm{static}} = \mathcal{E}_{v}\left(\mathbb{O}_{t,k}\right), 
\label{eq:imageencoder}
\end{equation}
where $k$ denotes the panoramic perspective index, $\mathcal{E}_{v}$ denotes the pre-trained visual encoder. These features are treated as static, since they are all captured at a single instant. Building upon this, our approach explicitly incorporates the dynamic features of nearby people. The agent then reasons over the structure formed by the waypoint predictor and the representation from the feature encoder, and commits to one node as its next action.

\textbf{Human Detection.} When the agent reaches a new waypoint, it needs to wait for the waypoint predictor to estimate new navigable areas. We leverage this property of waypoint-based navigation to design a short temporal observation strategy. 
During this brief pause, a human detector $\mathcal{D}$ analyzes the panorama $\mathcal{V}_t^{\mathrm{rgb}}$ and returns a set of bounding boxes $\mathcal{H}_t^{\mathrm{det}}$ for humans. 
If no human is detected, the agent will skip the remaining steps and proceed normally, selecting its next waypoint from $\mathcal{W}_{t}$ according to the navigation policy $w_{t+1} = \pi(\mathbb{O}_{<=t},\,\mathcal{I},\,\mathcal{H}_{<=t})$, where $\mathcal{H}_{<=t}$ represents the features collected related to humans up to this point.
When humans are detected, the agent continues to use this brief pause to acquire an observational sequence $\mathcal{S} = \langle \mathbb{O}_\tau \rangle_{\tau=t}^{t+m-1}$ to obtain dynamic information, where $m$ denotes the gathered timesteps.

\textbf{Human Reasoning.}
Upon detecting nearby humans, the agent reorganizes the temporal observation sequence $\mathcal{S}$ into individual-centric trajectories $\mathcal{P}_j^{1:m}$. To achieve holistic human understanding, we process each trajectory through two specialized streams. The geometric encoder captures spatio-temporal motion patterns for future occupancy reasoning, while the semantic encoder leverages a VLM to infer social context and human intent. Formally,
\begin{equation}
\mathcal{F}_{t,k,j}^{\text{geo}} = \mathcal{E}_{g}\!\left(\mathcal{P}_j^{1:m}\right),
\label{eq:geo}
\end{equation}
\begin{equation}
\mathcal{F}_{t,k,j}^{\text{sem}} = \mathcal{E}_{s}\!\left(\mathcal{P}_j^{1:m}\right),
\label{eq:sem}
\end{equation}
where $\mathcal{E}_{g}$ and $\mathcal{E}_{s}$ denote the geometric and semantic encoders, $t$ denotes the navigation timestep, $k$ denotes the panoramic perspective index, and $j$ denotes the detected human index. Detailed architectures are provided in Sec.~\ref{sec:Geometric Reasoning} and Sec.~\ref{sec:Semantic Reasoning}.


\textbf{Topological Feature Fusion.} 
To enable the navigation policy to reason over human dynamics, we inject the synchronized human features into the agent's topological representation. Specifically, we use a fusion layer to aggregate the static scene features $\mathcal{F}_{t,k}^{\text{static}}$ with the mean dynamic representations of all $J$ humans associated with a specific waypoint node:
\begin{equation}
\mathcal{F}_{t,k}^{\text{fused}} = \mathrm{MLP}\left(\mathcal{F}_{t,k}^{\text{static}},\frac{1}{J}\sum_{j=1}^{J} (\mathcal{F}_{t,k,j}^{\text{geo}}+\mathcal{F}_{t,k,j}^{\text{sem}})\right),
\label{eq:fusion}
\end{equation}
This human-aware topological node feature $\mathcal{F}_{t,k}^{\text{fused}}$ allows the Cross-modal Graph Transformer to align the global linguistic instruction $\mathcal{I}$ with localized, human-centric constraints. The final action $a_{t+1}$ is selected by evaluating candidate waypoints through a Feed-Forward Network (FFN), ensuring the trajectory is both goal-directed and socially appropriate.
\begin{equation}
a_{t+1} = \mathrm{FFN}(\mathrm{GASA}\left(\mathcal{F}_{t}^{\mathrm{fused}},\mathcal{W}_{t},\mathcal{I}\right)).
\label{eq:final}
\end{equation}


Having described how human-centric reasoning is integrated into the navigation process, we next detail the construction of the geometric and semantic human-centric features in Sec.~\ref{sec:Geometric Reasoning} and Sec.~\ref{sec:Semantic Reasoning}, followed by the social distance loss in Sec.~\ref{sec:Social Distance Loss}.

\subsection{Geometric Reasoning}
\label{sec:Geometric Reasoning}

Following the integrated reasoning pipeline described in Sec.~\ref{sec:Human Reasoning}, the Geometric Reasoning module specifically addresses the geometric deficit by forecasting fine-grained pose and coarse-grained trajectory.
For human-centric geometric reasoning, we decompose human motion into two complementary cues: body posture and movement trajectory, as shown in Fig.~\ref{fig:geo}. Body posture provides fine-grained cues about the activity in which a person is currently engaged, while movement trajectory offers coarse-grained evidence about where the person may move next. Consequently, our module is designed to interpret human behavior precisely through these two facets via estimation and prediction.

\textbf{Observed Geometric State Estimation.} For the collected sequence $\mathcal{S}$, when we organize the information by human index $j$, we extract the keypoints data of the human body to estimate the pose:
\begin{equation}
\mathcal{P}_{j}^{\mathrm{pose}} = \left\{ \mathbf{K}_{j}^{m} \right\}_{m=1}^{M}, 
\quad 
\mathbf{K}_{j}^{m} = \left\{ \mathbf{k}_{i,m}^{\mathrm{pose}} \right\}_{i=1}^{17} \in \mathbb{R}^{17 \times 3},
\label{eq:pose}
\end{equation}
where $\mathbf{k}_{i,m}^{\mathrm{pose}} = (x_{i,m}^{\mathrm{kp}}, y_{i,m}^{\mathrm{kp}}, c_{i,m}^{\mathrm{conf}})$ denotes the 2D joint position and detection confidence of the $i$-th keypoint at frame $m$, and $M$ denotes the observation length.

Simultaneously, we estimate the relative human position $\mathcal{P}_{j}^{\mathrm{traj}}$ by back-projecting the depth value at the human bounding-box center into 3D space using the standard pinhole camera model and the camera intrinsic matrix~\cite{hartley2003multiple}. The resulting 3D point is then transformed from the camera frame to the agent-centric frame according to the current agent heading, and its horizontal coordinates are taken as the trajectory cue for downstream geometric reasoning.

\textbf{Future Geometric State Forecasting.}
Once the data are reorganized, the agent predicts the future pose and trajectory of each person.
As shown in Fig.~\ref{fig:geo}, both the pose and the trajectory predictors share an underlying architecture composed of two core components. 
An LSTM-based encoder distills salient motion patterns from sequential inputs. Furthermore, a lightweight decoder propagates state predictions frame-by-frame. 
The features output by the encoder module will be fused and embedded to the graph representation for subsequent navigation module.

Crucially, the pose and trajectory prediction losses are computed on every step and their gradients back-propagated to fine-tune the model while the agent navigates on the train set.
Concretely, the pose prediction loss combines coordinate and confidence accuracy:
\begin{equation}
\label{eq:pose_loss}
\mathcal{L}_{\text{pose}} = \frac{1}{N}\sum_{i=1}^{N} \left( \| \mathbf{k}_i^{\text{pred}} - \mathbf{k}_i^{\text{gt}} \|_2^2 + \gamma_1 (c_i^{\text{pred}} - c_i^{\text{gt}})^2 \right),
\end{equation}
with $\gamma_1$ regulating confidence importance. And trajectory prediction integrates position and velocity constraints:
\begin{equation}
\label{eq:traj_loss}
\mathcal{L}_{\text{traj}} = \frac{1}{T}\sum_{t=1}^{T} \left( \| \mathbf{t}_t^{\text{pred}} - \mathbf{t}_t^{\text{gt}} \|_2^2 + \gamma_2 \| \mathbf{v}_t^{\text{pred}} - \mathbf{v}_t^{\text{gt}} \|_2^2 \right),
\end{equation}
as training progresses, the relative weights of these two losses in the total objective are progressively annealed. 

\textbf{Future-oriented Geometric Feature Construction.}
The forecasting task is designed not only to predict future human states, but also to learn intermediate representations that contain future-aware motion information. We therefore use the hidden representations from the trajectory and pose forecasting branches as the geometric feature $\mathcal{F}^{\mathrm{geo}}_{t,k,j}$ for human $j$. Since these representations are learned under explicit future prediction supervision, they are encouraged to capture imminent motion tendencies and potential future occupancy. The resulting geometric feature is then passed to the topological feature fusion module in Sec.~\ref{sec:Human Reasoning}, where it is combined with semantic human features and static scene features for downstream waypoint selection. The effectiveness of this future-oriented design is evaluated in the ablation studies.


\subsection{Semantic Reasoning}
\label{sec:Semantic Reasoning}

\begin{figure*}[t]
    \centering
    \includegraphics[width=0.9\linewidth]{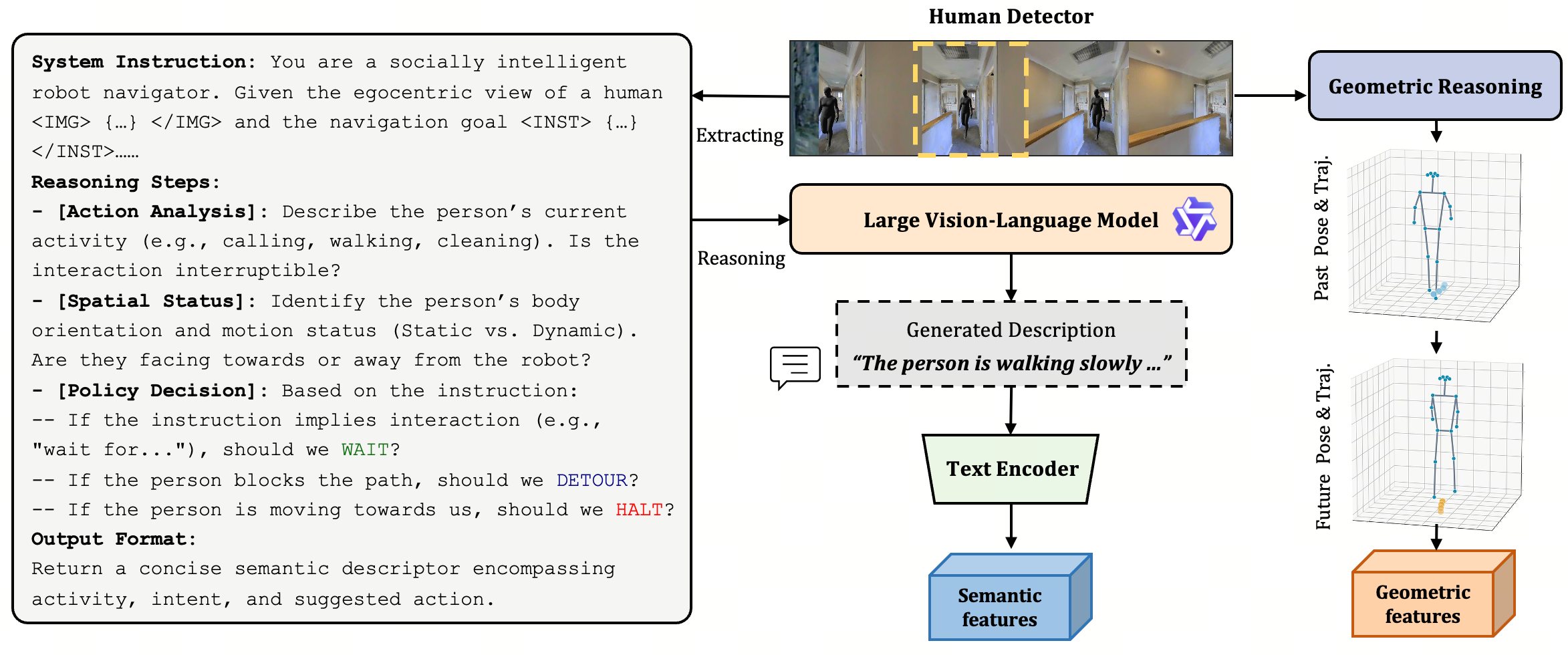}
    \caption{Illustration of human-centric semantic-geometric reasoning. Given the human detected in the image sequence, the agent performs semantic reasoning to infer activity, intent, and navigation-relevant social context, while geometric reasoning captures trajectory and pose cues, including their temporal evolution. These complementary cues provide structured human-aware representations for downstream navigation.}
    \label{fig:sem}
\end{figure*}

While geometric forecasting ensures the basic understanding and physical safety, it lacks the capability to interpret the high-level semantics of human activities, which is essential for grounding instructions such as \textit{``wait for the person calling''}. To bridge this gap, we design a Semantic Reasoning stream that leverages the zero-shot reasoning capability of Large Vision-Language Models.

\textbf{Visual-to-Linguistic Interpretation.}
For each detected human $j$ at timestep $t$, we first extract and crop the corresponding perspective view from the panoramic observation $V_t^{rgb}$ based on the human detector. This image $\mathcal{I}_{t,k,j}$ serves as the visual input to the semantic branch. As illustrated in Fig.~\ref{fig:sem}, the goal of this branch is to infer human activity, intent, and navigation-relevant social context, thereby complementing the geometric cues with higher-level semantics. To guide the Vision-Language Model toward navigation-relevant reasoning rather than generic captioning, we formulate a task-specific prompt $\mathcal{Q}$. Conditioned on $\mathcal{I}_{t,k,j}$ and $\mathcal{Q}$, the VLM generates a natural-language description $\mathcal{T}_{t,k,j}$:
\begin{equation}
\mathcal{T}_{t,k,j} = \text{VLM}(I_{t,k,j}, \mathcal{Q}).
\end{equation}
For instance, the model might output \textit{``A person is standing still and sorting clothes''} or \textit{``A person is walking rapidly towards the corridor''}. This linguistic representation explicitly captures the high-level semantic intent that the geometric reasoning module fails to convey.

\textbf{Semantic Feature Encoding.}
To integrate this linguistic insight into the navigation policy, we instantiate the semantic encoder $\mathcal{E}_s$ with a pre-trained text encoder (i.e. CLIP~\cite{radford2021learning}), which maps the generated description $\mathcal{T}_{t,k,j}$ into a high-dimensional semantic representation:
\begin{equation}
\mathcal{F}_{t,k,j}^{\mathrm{sem}} = \mathcal{E}_{s}\!\left(\mathcal{T}_{t,k,j}\right).
\label{eq:semantic_feature}
\end{equation}
Here, $\mathcal{E}_{s}$ denotes the text encoding stage of the semantic branch. Unlike the geometric feature $\mathcal{F}_{t,k,j}^{\mathrm{geo}}$, which encodes \textit{how} a person moves, $\mathcal{F}_{t,k,j}^{\mathrm{sem}}$ captures \textit{why} the person acts, providing crucial context for instruction alignment. Finally, this semantic representation is fused with the geometric features, enabling the Cross-modal Graph Transformer to reason jointly over physical constraints and social context.

\subsection{Social Distance Losses}
\label{sec:Social Distance Loss}
Beyond the reasoning modules, we design a unified Social Distance Loss that discourages the planner from outputting any trajectory that would intrude upon space already occupied by humans, thereby ensuring collision-free motion and safe human-robot interaction. It is enforced through two complementary mechanisms:

\textbf{Actual Collision Loss.} During navigation, each collision is penalized by:
\begin{equation}
\label{eq:coll_loss}
\mathcal{L}_{\text{coll}} = \lambda_c \sum_{i\in\mathcal{C}} \delta_i,
\end{equation}
where $\mathcal{C}$ is the set of collision events and $\delta_i$ is the penalty coefficient (empirically set to 3.0).
    
\textbf{Proximity Avoidance Loss.} For humans detected within safety radius $r_s=1.0\,\text{m}$, we compute a repulsive loss that grows as the robot approaches the human:
\begin{equation}
\label{eq:prox_loss}
\mathcal{L}_{\text{prox}} = \lambda_p \sum_{j} \frac{\phi(\theta_j)}{ \max(\|\mathbf{d}_j\|_2^2, \epsilon_p) },
\end{equation}
where $\mathbf{d}_j = (d_x^j, d_y^j)$ is the relative position vector, $\phi(\cdot)$ is the front-facing penalty weighting function ($[0.25,1.0]$), and $\epsilon_p=0.0625\,\text{m}^2$ prevents divergence.
    
Finally, combining the auxiliary forecasting losses in Eqs.~\eqref{eq:pose_loss} and~\eqref{eq:traj_loss}, the social safety losses in Eqs.~\eqref{eq:coll_loss} and~\eqref{eq:prox_loss}, and the standard navigation imitation objective, the overall training objective is formulated as:
\begin{equation}
\mathcal{L}_{\text{total}} =
\mathcal{L}_{\text{pose}}
+
\mathcal{L}_{\text{traj}}
+
\mathcal{L}_{\text{coll}}
+
\mathcal{L}_{\text{prox}}
+
\mathcal{L}_{\text{nav}},
\label{eq:total_loss}
\end{equation}
where $\mathcal{L}_{\text{nav}} = -\mathbb{E}[\log P(a^*|s)]$ denotes the standard cross-entropy loss for navigation action prediction.

\section{Experiments}

\subsection{Experiment Setup}
\subsubsection{Dataset}
The HA-VLNCE dataset integrates 486 SMPL-format 3D human motion models spanning 172 daily activity categories (including running, climbing, and phone conversations, totaling 58,320 frame sequences), annotates 910 dynamic human instances across 90 scenes (with up to 10 humans per scene), and extends 16,844 human-centric instructions averaging 15 words in length, comprising a training set of 10,819 instructions for learning human dynamic interactions and path planning, along with validation sets containing 778 seen validation set for evaluating generalization in familiar environments and 1,839 unseen validation set testing adaptability in novel scenarios, collectively establishing a continuous human-aware VLN benchmark for assessing agent performance in human-populated environments~\cite{dong2025ha}.


\subsubsection{Evaluation Metrics}
Following HA-VLNCE~\cite{dong2025ha}, we evaluate the agent using navigation and human-safety metrics, including Navigation Error (NE), Success Rate (SR), Total Collision Rate (TCR), and Collision Rate (CR). NE measures the average distance between the agent's final position and the target, while SR measures the proportion of episodes successfully completed without collision. TCR reflects the overall frequency of human-related collisions throughout navigation, and CR measures the proportion of episodes that contain at least one human-related collision. Since this work focuses on human-populated navigation, our main comparison emphasizes SR, TCR, and CR, which directly reflect task completion and social safety in dynamic human environments. NE is included in the evaluation protocol for completeness, but the main tables prioritize safety-related metrics.

\begin{table*}[t]
\centering
\caption{\textbf{Comparison with state-of-the-art methods on the HA-VLNCE benchmark.} 
Best results are highlighted in bold and second-best results are underlined. 
For CR and SR, we additionally report the relative improvement of our method over the second-best baseline.}
\label{tab:sota_comparison}
\setlength{\tabcolsep}{6.0pt}
\renewcommand{\arraystretch}{1.15}
\begin{tabularx}{\textwidth}{@{} l @{\hspace{1.0em}} *{6}{>{\centering\arraybackslash}X} @{}}
\toprule[1.2pt]
\multirow{2}{*}{\textbf{Models}}
& \multicolumn{3}{c}{\textbf{Validation Seen}}
& \multicolumn{3}{c}{\textbf{Validation Unseen}} \\
\cmidrule(l{0.8em}r{0.8em}){2-4}
\cmidrule(l{0.8em}r{0.8em}){5-7}
& \textbf{TCR}$\downarrow$ & \textbf{CR}$\downarrow$ & \textbf{SR}$\uparrow$
& \textbf{TCR}$\downarrow$ & \textbf{CR}$\downarrow$ & \textbf{SR}$\uparrow$ \\
\midrule
HA-VLN-CMA~\cite{dong2025ha}
& 63.09 & 0.77 & 0.05
& 47.06 & 0.77 & 0.07 \\

HA-VLN-CMA-DA~\cite{dong2025ha}
& 17.45 & 0.61 & 0.17
& 27.25 & 0.69 & 0.09 \\

HA-VLN-VL~\cite{dong2025ha}
& 4.44 & 0.52 & 0.20
& 6.63 & 0.59 & 0.14 \\

LAW-VLNCE~\cite{raychaudhuri2021language}
& 4.31 & 0.54 & 0.21
& 5.88 & 0.65 & 0.15 \\

DUET~\cite{chen2022think}
& 4.18 & 0.48 & 0.22
& 5.74 & 0.63 & 0.16 \\

ETPNav~\cite{an2024etpnav}
& 4.07 & \underline{0.43} & 0.24
& 6.94 & 0.58 & 0.17 \\

GridMM~\cite{wang2023gridmm}
& 3.92 & 0.45 & 0.24
& 5.76 & 0.59 & 0.18 \\

BEVBert~\cite{an2022bevbert}
& \underline{3.64} & 0.46 & \underline{0.27}
& \textbf{4.71} & \underline{0.55} & \underline{0.21} \\

\rowcolor{gray!20}
\textbf{HCSG (Ours)}
& \textbf{3.63}
& \textbf{0.34}\,\scriptsize{($\downarrow$20.9\%)}
& \textbf{0.29}\,\scriptsize{($\uparrow$7.4\%)}
& \underline{5.02}
& \textbf{0.36}\,\scriptsize{($\downarrow$34.5\%)}
& \textbf{0.24}\,\scriptsize{($\uparrow$14.3\%)} \\
\bottomrule[1.2pt]
\end{tabularx}
\end{table*}

\begin{figure}[t!]
    \centering
    \includegraphics[width=1\linewidth]{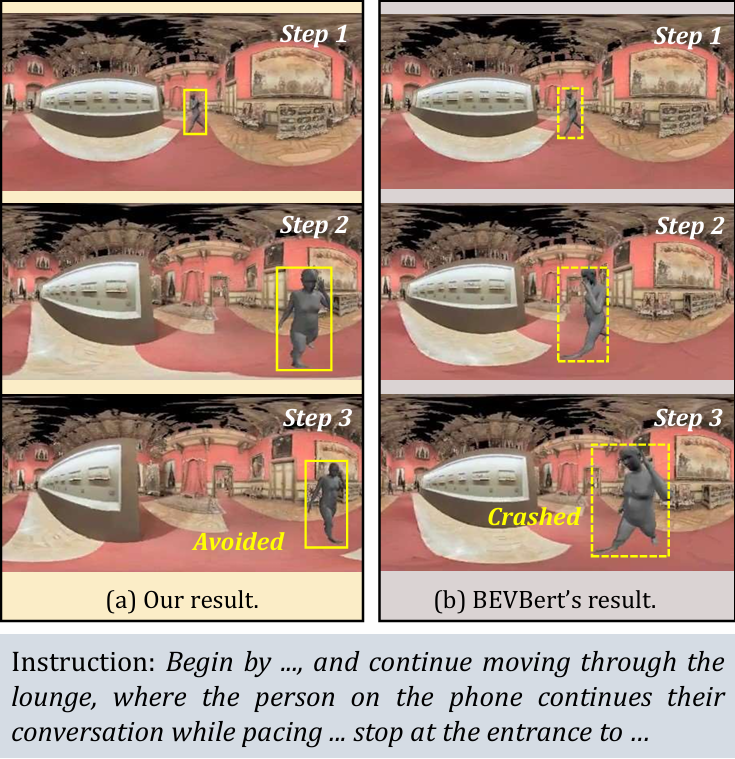}
    \caption{Visualization of results on HA‑VLNCE benchmark. 
    (a) shows that our model avoided the person and continued to follow the instruction, while the SOTA model (BEVBert) in (b) collided with the person.
}
    \label{fig:Case Study2}
\end{figure}

\begin{figure}[t!]
    \centering
    \includegraphics[width=1\linewidth]{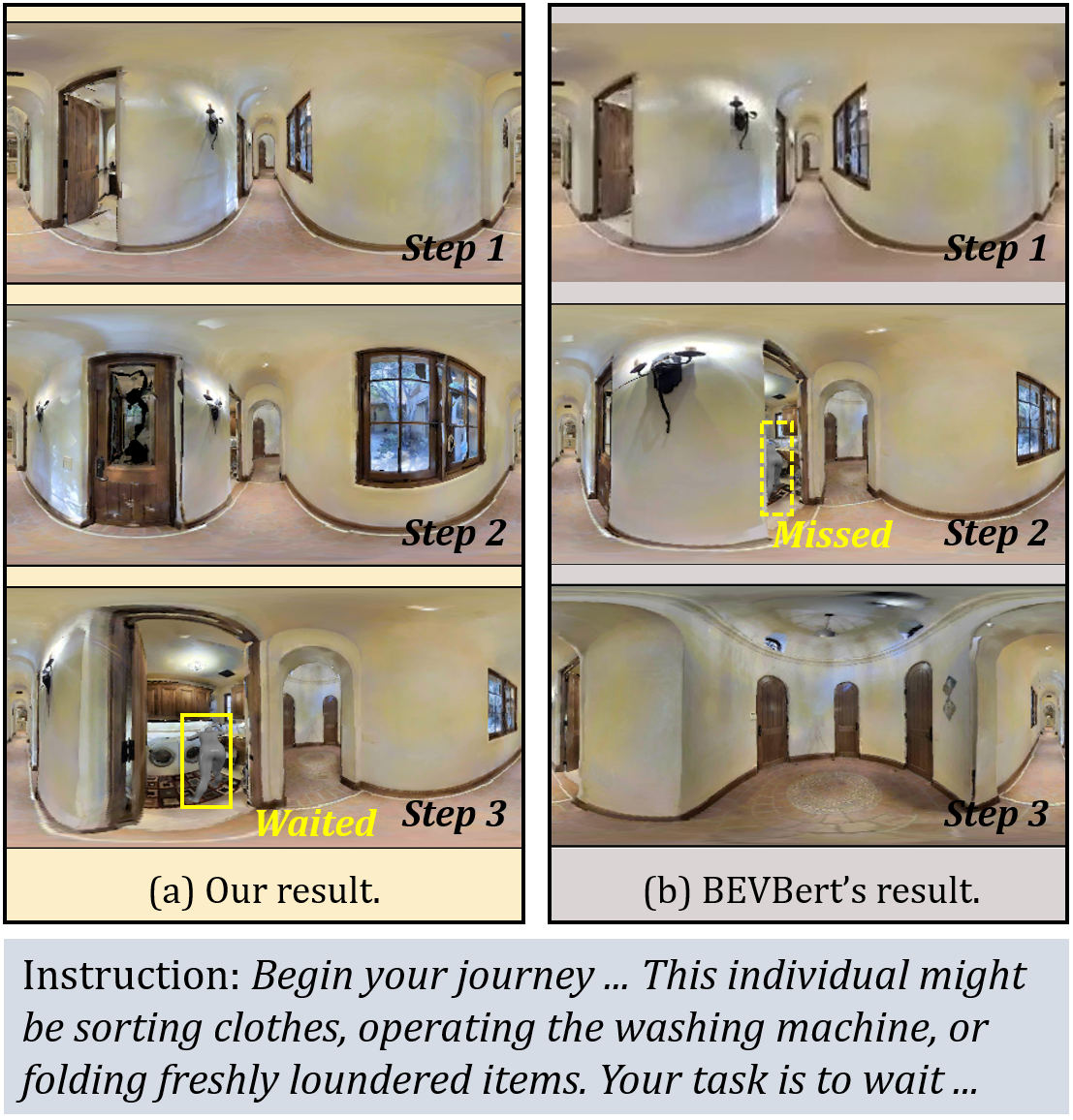}
    \caption{Visualization of results on HA‑VLNCE benchmark. 
    (a) shows that our model recognized the human behavior in the language description and successfully completed the VLN task, while the SOTA model (BEVBert) in (b) missed this person.
}
    \label{fig:Case Study}
\end{figure}

\subsubsection{Implementation Details}

All experiments are conducted using PyTorch framework on two NVIDIA A6000 GPUs. We pre-train the model in static environments~\cite{krantz2020beyond} for 100{,}000 iterations with a batch size of~64 and a learning rate of $5\times10^{-5}$, employing the AdamW optimizer. During fine-tuning, the agent interacts with the environments online via the Habitat Simulator~\cite{habitat19iccv}. At each waypoint, we apply YOLO‑Pose~\cite{maji2022yolo} to perform human detection; once a person is detected, the agent collects 6 frames of information and predicts the trajectories and poses for the next 3 frames. We employ Qwen3-VL-2B-Instruct~\cite{qwen3technicalreport} to reason human motion. 
Fine-tuning is performed for 15{,}000 iterations with a batch size of~8 and a learning rate of $1\times10^{-5}$.

\subsection{Comparisons with State-of-the-art}

As evidenced in Table~\ref{tab:sota_comparison}, HCSG establishes a new state-of-the-art on the HA-VLNCE dataset. A critical observation from the baseline methods is the inherent trade-off between navigation success and safety; traditional agents tend to improve goal-reaching performance without explicitly controlling human-related collisions. However, our method effectively overcomes this limitation. Under the challenging validation-unseen split, HCSG reduces the critical Collision Rate (CR) to 0.36, yielding a significant reduction of 34.5\% compared to BEVBert and 37.9\% compared to ETPNav. Concurrently, our agent attains a Success Rate (SR) of 0.24, surpassing BEVBert (0.21) and ETPNav (0.17). This simultaneous improvement in both success and safety metrics demonstrates that explicitly modeling human behaviors allows the agent to navigate aggressively yet safely, rather than simply stopping or colliding with pedestrians when encountering pedestrians.

Fig.~\ref{fig:Case Study2} and Fig.~\ref{fig:Case Study} provide qualitative visualizations of our model's performance on the HA-VLNCE benchmark. In Fig.~\ref{fig:Case Study}, the instruction requires the agent to wait for a person sorting clothes. Traditional methods fail to capture this semantic cue and attempt to bypass the person immediately, leading to a collision or task failure. In contrast, our HCSG correctly interprets the activity description, halts at a safe distance, and resumes navigation once the path is clear. Similarly, Fig.~\ref{fig:Case Study2} demonstrates a dynamic avoidance scenario where the agent anticipates the trajectory of a pacing person. These visualizations qualitatively validate that our framework succeeds not only by reacting to obstacles but by understanding the semantic intent and future geometry of human motion.

\subsection{Ablation Studies}
\subsubsection{Significance of Geometric Reasoning}

\begin{table}[t]
\centering
\setlength{\tabcolsep}{6pt}
\caption{\textbf{Ablation study on geometric reasoning.}}
\label{tab:Importance of Human Understanding Fusion.}
\scriptsize{%
\renewcommand{\arraystretch}{1.15}
\begin{tabularx}{\linewidth}{cc|>{\centering\arraybackslash}X>{\centering\arraybackslash}X>{\centering\arraybackslash}X|>{\centering\arraybackslash}X>{\centering\arraybackslash}X>{\centering\arraybackslash}X}
\toprule[1.2pt]
\multirow{2}{*}{\textbf{Pose}} & \multirow{2}{*}{\textbf{Trajectory}} 
& \multicolumn{3}{c}{\textbf{Validation Seen}} 
& \multicolumn{3}{c}{\textbf{Validation Unseen}} \\ 
\cmidrule(lr){3-5} \cmidrule(lr){6-8}
& & \textbf{TCR$\downarrow$} & \textbf{CR$\downarrow$} & \textbf{SR$\uparrow$} 
& \textbf{TCR$\downarrow$} & \textbf{CR$\downarrow$} & \textbf{SR$\uparrow$} \\ 
\midrule
\xmark & \xmark & 3.98 & 0.39 & 0.26 & 6.09 & 0.48 & 0.20 \\ 
\cmark & \xmark & 3.91 & 0.38 & \textbf{0.29} & 5.92 & 0.47 & 0.22 \\ 
\xmark & \cmark & \textbf{3.72} & \textbf{0.36} & 0.27 & 5.31 & 0.41 & 0.22 \\ 
\cmark & \cmark & \textbf{3.72} & \textbf{0.36} & 0.28 & \textbf{5.21} & \textbf{0.40} & \textbf{0.23} \\ 
\bottomrule[1.2pt]
\end{tabularx}%
}
\end{table}

\begin{table}[!t]
\centering
\setlength{\tabcolsep}{6pt}
\caption{\textbf{Ablation study on VLM-based semantic reasoning.}}
\label{tab:vlm_ablation}
\scriptsize{%
\renewcommand{\arraystretch}{1.18}
\begin{tabularx}{\linewidth}{c|>{\centering\arraybackslash}X>{\centering\arraybackslash}X>{\centering\arraybackslash}X|>{\centering\arraybackslash}X>{\centering\arraybackslash}X>{\centering\arraybackslash}X}
\toprule[1.2pt]
\multirow{2}{*}{\textbf{Setting}} 
& \multicolumn{3}{c}{\textbf{Validation Seen}} 
& \multicolumn{3}{c}{\textbf{Validation Unseen}} \\ 
\cmidrule(l{0.8em}r{0.8em}){2-4} 
\cmidrule(l{0.8em}r{0.8em}){5-7} 
& \textbf{TCR$\downarrow$} & \textbf{CR$\downarrow$} & \textbf{SR$\uparrow$} 
& \textbf{TCR$\downarrow$} & \textbf{CR$\downarrow$} & \textbf{SR$\uparrow$} \\ 
\midrule
w/o VLM 
& 3.72 & 0.36 & 0.28 
& 5.21 & 0.40 & 0.23 \\[-2pt]
\textbf{w VLM} 
& \textbf{3.66} & \textbf{0.34} & \textbf{0.29} 
& \textbf{5.02} & \textbf{0.36} & \textbf{0.24} \\ 
\bottomrule[1.2pt]
\end{tabularx}%
}
\end{table}

\begin{table}[t]
\centering
\setlength{\tabcolsep}{6pt}
\caption{\textbf{Ablation study on future vs. past-oriented feature.}}
\label{tab:FT}
\scriptsize{%
\renewcommand{\arraystretch}{1.18}
\begin{tabularx}{\linewidth}{c|>{\centering\arraybackslash}X>{\centering\arraybackslash}X>{\centering\arraybackslash}X|>{\centering\arraybackslash}X>{\centering\arraybackslash}X>{\centering\arraybackslash}X}
\toprule[1.2pt]
\multirow{2}{*}{\textbf{Feature}} 
& \multicolumn{3}{c}{\textbf{Validation Seen}} 
& \multicolumn{3}{c}{\textbf{Validation Unseen}} \\ 
\cmidrule(l{0.8em}r{0.8em}){2-4} 
\cmidrule(l{0.8em}r{0.8em}){5-7} 
& \textbf{TCR$\downarrow$} & \textbf{CR$\downarrow$} & \textbf{SR$\uparrow$} 
& \textbf{TCR$\downarrow$} & \textbf{CR$\downarrow$} & \textbf{SR$\uparrow$} \\ 
\midrule
Past-oriented 
& 3.81 & 0.37 & 0.27 
& 5.25 & 0.41 & 0.21 \\[-2pt]
\textbf{Future-oriented} 
& \textbf{3.72} & \textbf{0.36} & \textbf{0.28} 
& \textbf{5.21} & \textbf{0.40} & \textbf{0.23} \\ 
\bottomrule[1.2pt]
\end{tabularx}%
}
\end{table}

\begin{table}[!t]
\centering
\setlength{\tabcolsep}{6pt}
\caption{\textbf{Ablation study on fine-tuning vs. fixed-parameter design.}}
\label{tab:Comparison of Embedded/External Design.}
\scriptsize{%
\renewcommand{\arraystretch}{1.18}
\begin{tabularx}{\linewidth}{c|>{\centering\arraybackslash}X>{\centering\arraybackslash}X>{\centering\arraybackslash}X|>{\centering\arraybackslash}X>{\centering\arraybackslash}X>{\centering\arraybackslash}X}
\toprule[1.2pt]
\multirow{2}{*}{\textbf{Setting}} 
& \multicolumn{3}{c}{\textbf{Validation Seen}} 
& \multicolumn{3}{c}{\textbf{Validation Unseen}} \\ 
\cmidrule(l{0.8em}r{0.8em}){2-4} 
\cmidrule(l{0.8em}r{0.8em}){5-7} 
& \textbf{TCR$\downarrow$} & \textbf{CR$\downarrow$} & \textbf{SR$\uparrow$} 
& \textbf{TCR$\downarrow$} & \textbf{CR$\downarrow$} & \textbf{SR$\uparrow$} \\ 
\midrule
Fixed-parameter 
& 3.86 & 0.39 & 0.27 
& 5.27 & 0.42 & 0.22 \\[-2pt]
\textbf{Fine-tuning} 
& \textbf{3.72} & \textbf{0.36} & \textbf{0.28} 
& \textbf{5.21} & \textbf{0.40} & \textbf{0.23} \\ 
\bottomrule[1.2pt]
\end{tabularx}%
}
\end{table}

\begin{table}[t]
\centering
\setlength{\tabcolsep}{6pt}
\caption{\textbf{Ablation on Social Distance Loss.}}
\label{tab:Importance of Social Distance Loss.}
\scriptsize{%
\renewcommand{\arraystretch}{1.15}
\begin{tabularx}{\linewidth}{cc|>{\centering\arraybackslash}X>{\centering\arraybackslash}X>{\centering\arraybackslash}X|>{\centering\arraybackslash}X>{\centering\arraybackslash}X>{\centering\arraybackslash}X}
\toprule[1.2pt]
\multirow{2}{*}{\textbf{$\mathcal{L}_{\text{coll}}$}} 
& \multirow{2}{*}{\textbf{$\mathcal{L}_{\text{prox}}$}} 
& \multicolumn{3}{c}{\textbf{Validation Seen}} 
& \multicolumn{3}{c}{\textbf{Validation Unseen}} \\ 
\cmidrule(lr){3-5} \cmidrule(lr){6-8}
& & \textbf{TCR$\downarrow$} & \textbf{CR$\downarrow$} & \textbf{SR$\uparrow$} 
& \textbf{TCR$\downarrow$} & \textbf{CR$\downarrow$} & \textbf{SR$\uparrow$} \\ 
\midrule 
\xmark & \xmark 
& 4.06 & 0.42 & 0.26 
& 6.73 & 0.55 & 0.20 \\ 
\cmark & \xmark 
& 4.02 & 0.41 & 0.26 
& 6.35 & 0.52 & 0.21 \\ 
\xmark & \cmark 
& 3.99 & 0.39 & 0.27 
& 6.13 & 0.50 & 0.21 \\ 
\cmark & \cmark 
& \textbf{3.72} & \textbf{0.36} & \textbf{0.28} 
& \textbf{5.21} & \textbf{0.40} & \textbf{0.23} \\ 
\bottomrule[1.2pt]
\end{tabularx}%
}
\end{table}

\begin{table}[!t]
\centering
\setlength{\tabcolsep}{6pt}
\caption{\textbf{Ablation on RGB/depth inputs.}}
\label{tab:agent_performance}
\scriptsize{%
\renewcommand{\arraystretch}{1.15}
\begin{tabularx}{\linewidth}{cc|>{\centering\arraybackslash}X>{\centering\arraybackslash}X>{\centering\arraybackslash}X|>{\centering\arraybackslash}X>{\centering\arraybackslash}X>{\centering\arraybackslash}X}
\toprule[1.2pt]
\multirow{2}{*}{\textbf{RGB}} 
& \multirow{2}{*}{\textbf{Depth}} 
& \multicolumn{3}{c}{\textbf{Validation Seen}} 
& \multicolumn{3}{c}{\textbf{Validation Unseen}} \\
\cmidrule(lr){3-5} \cmidrule(lr){6-8}
& & \textbf{TCR$\downarrow$} & \textbf{CR$\downarrow$} & \textbf{SR$\uparrow$}
& \textbf{TCR$\downarrow$} & \textbf{CR$\downarrow$} & \textbf{SR$\uparrow$} \\
\midrule[0.6pt]
\cmark & \xmark 
& 5.93 & 0.51 & 0.20 
& 7.31 & 0.56 & 0.15 \\
\xmark & \cmark 
& 4.48 & 0.40 & 0.26 
& 7.02 & 0.46 & 0.21 \\
\cmark & \cmark 
& \textbf{3.72} & \textbf{0.36} & \textbf{0.28} 
& \textbf{5.21} & \textbf{0.40} & \textbf{0.23} \\
\bottomrule[1.2pt]
\end{tabularx}%
}
\end{table}

We first analyze the impact of the explicit geometric reasoning components, as detailed in Table~\ref{tab:Importance of Human Understanding Fusion.}. The baseline model without any specific human understanding modules yields a high collision rate of 0.48 in unseen environments. Incorporating trajectory features alone lowers the collision rate significantly to 0.41. This result indicates that predicting future occupancy helps the agent proactively plan paths that avoid dynamic obstacles. On the other hand, adding pose features alone results in a noticeable improvement in Success Rate, increasing from 0.20 to 0.22. This suggests that pose information serves as a fine-grained semantic signal, helping the agent distinguish between different human activities relevant to the instruction. The synergistic fusion of both pose and trajectory features achieves the optimal performance with a CR of 0.40 and SR of 0.23. This confirms that geometric constraints and behavioral semantics are complementary; the former ensures physical safety while the latter aids in task grounding.

\subsubsection{Impact of VLM-based Semantic Reasoning}
To further verify the necessity of the Semantic Stream, we conduct an ablation study on the VLM component, as shown in Table~\ref{tab:vlm_ablation}. The model without VLM support relies solely on geometric cues for human avoidance. While it achieves reasonable safety, the integration of VLM-based descriptions further optimizes performance. Specifically, the inclusion of VLM reduces the Collision Rate from 0.40 to 0.36 and improves the Success Rate from 0.23 to 0.24 in unseen environments. This improvement can be attributed to the VLM's capability to interpret complex social scenarios that geometric features cannot capture, such as identifying whether a person is stationary and interacting with an object or about to turn around. By converting visual human actions into linguistic descriptions, the VLM aligns the visual observations more effectively with the natural language instructions, leading to safer and more intelligent decision-making.

\subsubsection{Future vs. Past-oriented Feature}

In the previous section, we emphasize that we employ a future-oriented understanding approach, which involves having our agent predict the future state of humans to understand its current ongoing activities. We argue that this method is better than the past-oriented understanding approach of directly encoding the collected information, because only by knowing what a person will do in the future can it be said that one truly understands his behavior. Empirical results in Table~\ref{tab:FT} support this hypothesis. The future-oriented design outperforms the past-oriented counterpart in unseen scenarios, specifically increasing the Success Rate from 0.21 to 0.23 and decreasing the Collision Rate from 0.41 to 0.40. This confirms that predictive features provide more actionable guidance for navigation than historical encodings.

\subsubsection{Fine-tuning vs. Fixed-parameter Design} 

We further claim that the continuous fine-tuning strategy for the geometric reasoning module provides more adaptive human features that are more adaptable than fixed pre-trained features. To validate this, we compared our adaptive approach against a fixed-parameter baseline in Table~\ref{tab:Comparison of Embedded/External Design.}. The quantitative results confirm the advantage of adaptation. Specifically, the fine-tuning mechanism reduces the Collision Rate from 0.42 to 0.40 and improves the Success Rate from 0.22 to 0.23 in validation unseen environments. This improvement implies that the domain distribution of human behaviors in the navigation environment differs from the pre-training dataset. Consequently, training-time adaptation allows the motion forecasting module to adapt to specific movement patterns encountered during the episode, thereby providing more precise features for the navigation policy.

\subsubsection{Significance of Social Distance Loss}

\begin{figure}[!t]
    \centering
    \includegraphics[width=1\linewidth]{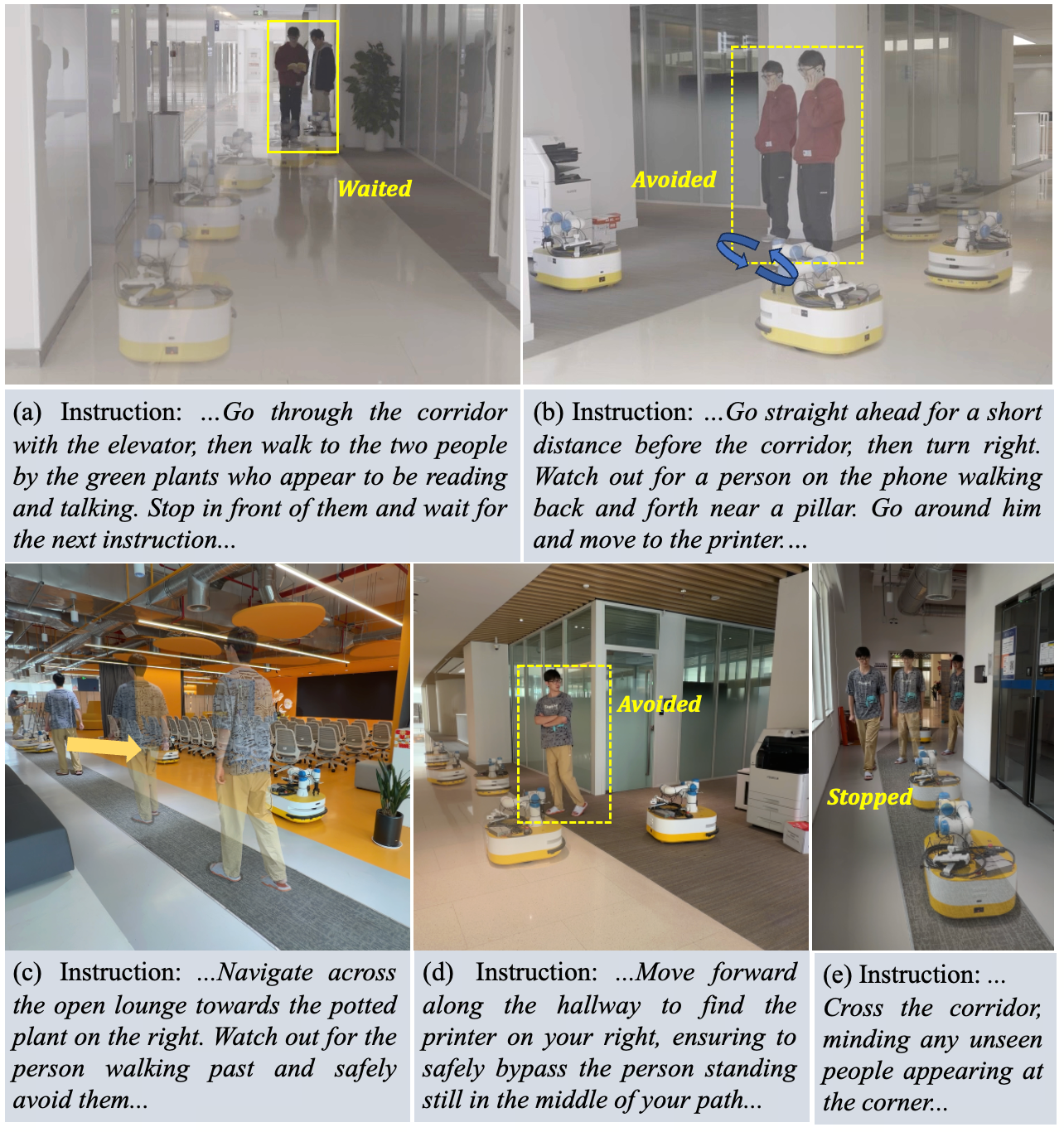}
    \caption{Qualitative results of real-world deployment on the NXROBO Leo platform. (a) Semantic reasoning enables the agent to approach and wait for interacting pedestrians. (b) Geometric forecasting allows proactive detour planning around a pacing person. (c) The agent safely avoids a pedestrian crossing its path in an open lounge. (d) It correctly identifies and bypasses a stationary human blocking a hallway. (e) It successfully anticipates and yields to an unseen pedestrian appearing from a blind corner.}
    \label{fig:real_world}
\end{figure}
As demonstrated in Table~\ref{tab:Importance of Social Distance Loss.}, the proposed dual loss mechanism exhibits complementary advantages in optimizing navigation safety. The collision penalty loss significantly reduces the Collision Rate from 0.55 to 0.52 in unseen environments. Concurrently, the proximity avoidance loss achieves a reduction in Total Collision Rate from 6.73 to 6.13. Crucially, their combined implementation creates a synergistic prevention-penalty mechanism that drives substantial further improvements. The combination lowers the Unseen Collision Rate to 0.40 and optimizes the Total Collision Rate to 5.21, validating the necessity of enforcing both immediate collision constraints and proactive spacing.

\subsubsection{Ablation on RGB/Depth inputs}

Table~\ref{tab:agent_performance} confirms the robustness of our model to different visual modalities. The model relying solely on RGB inputs suffers from a high Collision Rate of 0.56 and a low Success Rate of 0.15 in unseen scenes, as it lacks precise depth information to estimate distances to pedestrians. Integrating Depth information significantly alleviates this issue, lowering the Collision Rate to 0.46 and boosting the Success Rate to 0.21. However, the optimal performance is achieved only when both RGB and Depth inputs are utilized, which yields a Collision Rate of 0.40 and a Success Rate of 0.23. This indicates that RGB data provides essential semantic context for the VLM and pose estimation, while Depth data ensures accurate trajectory forecasting and safety constraint enforcement.

\begin{table}[t!]
\centering
\caption{Quantitative Results of Real-World Deployment. We compare the Success Rate of HCSG against BEVBert across the five distinct scenarios depicted in Fig.~\ref{fig:real_world}, with 10 trials per scenario.}
\label{tab:real_world}
\resizebox{\columnwidth}{!}{%
\begin{tabular}{l|cc}
\toprule
\multirow{2}{*}{\textbf{Real-World Scenarios}} & \multicolumn{2}{c}{\textbf{Success Rate $\uparrow$}} \\
\cmidrule{2-3}
 & \textbf{BEVBert~\cite{an2022bevbert}} & \textbf{HCSG (Ours)} \\
\midrule
Scenario (a): Stop in front of interacting pedestrians & 3/10 (30\%) & \textbf{9/10 (90\%)} \\
Scenario (b): Avoid pacing pedestrian on phone & 3/10 (30\%) & \textbf{7/10 (70\%)} \\
Scenario (c): Navigate to plant, avoid moving human & 5/10 (50\%) & \textbf{9/10 (90\%)} \\
Scenario (d): Bypass standing human in hallway & 6/10 (60\%) & \textbf{9/10 (90\%)} \\
Scenario (e): Traverse corridor, avoid corner pedestrian & 2/10 (20\%) & \textbf{7/10 (70\%)} \\
\midrule
\textbf{Overall Average} & 19/50 (38\%) & \textbf{41/50 (82\%)} \\
\bottomrule
\end{tabular}%
}
\end{table}

\subsection{Real-world Deployment}

To validate the sim-to-real transferability of HCSG, we conducted physical experiments using the NXROBO Leo mobile manipulator equipped with an RGB-D camera in a controlled office environment. We evaluated the system through both a systematic quantitative comparison and qualitative case demonstrations.

To systematically quantify the real-world capabilities, we established an evaluation benchmark comprising five distinct scenarios (depicted in Fig.~\ref{fig:real_world}) and conducted 10 physical trials per scenario. As summarized in Table~\ref{tab:real_world}, HCSG provides preliminary real-world validation, achieving an overall success rate of 82\%. In semantics-dependent tasks involving stationary humans (Scenarios a, d), BEVBert frequently failed by misclassifying interactive targets as mere obstacles or freezing, whereas our semantic stream ensured robust instruction grounding. Furthermore, in highly dynamic settings (Scenarios b, c, e), our geometric forecasting enabled proactive detour planning, yielding substantial improvements over the baseline's purely reactive approach and verifying the framework's effectiveness in real physical environments.

Beyond the quantitative metrics, Fig.~\ref{fig:real_world} provides qualitative visualizations that highlight the mechanisms behind our agent's robustness. First, the framework demonstrates accurate semantic understanding when handling stationary individuals. For instance, in Fig.~\ref{fig:real_world}(a), the Semantic Reasoning stream successfully identifies two talking pedestrians as the interaction targets, navigating to a socially appropriate distance to wait. Similarly, in Fig.~\ref{fig:real_world}(d), the agent correctly parses a stationary human blocking a hallway as a passive obstacle to be bypassed rather than an interaction target.

Moreover, our agent exhibits superior geometric foresight when encountering dynamic obstacles. In Fig.~\ref{fig:real_world}(b), the Geometric Reasoning module captures the temporal motion of a pacing pedestrian near a pillar and proactively plans a detour. This proactive capability extends seamlessly to diverse spatial layouts: the robot intelligently avoids a pedestrian crossing an open lounge (Fig.~\ref{fig:real_world}(c)) and safely handles a dynamic encounter with an unseen pedestrian suddenly appearing from a blind corner (Fig.~\ref{fig:real_world}(e)) without freezing or causing collisions.


\section{Conclusion}

In this work, we presented HCSG, a novel human-centric framework that endows VLN agents with explicit dual-stream human understanding. By combining geometric forecasting with semantic interpretation, our method enables the agent to reason about both future human motion and high-level human intent, moving beyond passive obstacle avoidance toward socially aware navigation. In addition, the proposed safety objective encourages the agent to maintain appropriate interaction distances around humans. Experimental results on the HA-VLNCE benchmark demonstrate that HCSG consistently improves both navigation success and collision reduction, highlighting the importance of explicit human-centric reasoning in dynamic indoor environments. Future work will focus on extending the current stop-and-wait design to a continuous streaming setting and on learning more adaptive social distance priors conditioned on richer human context. We believe these directions will further advance socially intelligent navigation in real-world human-robot environments.

\bibliographystyle{IEEEtran}  
\bibliography{refs}  

\end{document}